\documentclass[10pt,twocolumn,letterpaper]{article}

\usepackage{multirow}
\usepackage{graphicx}
\usepackage[pagenumbers]{cvpr} % To force page numbers, e.g. for an arXiv version

\expandafter\def\expandafter\normalsize\expandafter{%
    \normalsize%
    \setlength\abovedisplayskip{3pt}%
    \setlength\belowdisplayskip{3pt}%
    \setlength\abovedisplayshortskip{-3pt}%
    \setlength\belowdisplayshortskip{3pt}%
}

\usepackage{pifont}
\definecolor{tablegray}{gray}{0.9}
\definecolor{cvprblue}{rgb}{0.21,0.49,0.74}
\usepackage[pagebackref,breaklinks,colorlinks,allcolors=cvprblue]{hyperref}

\def\paperID{***}
\def\confName{CVPR}
\def\confYear{2025}

\title{Learning Video Representations without Natural Videos}

\author{Xueyang Yu\\
ShanghaiTech University\\
{\tt\small yuxy1@shanghaitech.edu.cn}
\and
Xinlei Chen\\
Meta AI\\
{\tt\small xinleic@meta.com}
\and
Yossi Gandelsman\\
University of California, Berkeley\\
{\tt\small yossi\_gandelsman@berkeley.edu}
}

\begin{document}
\maketitle

\begin{abstract}

We show that useful video representations can be learned from synthetic videos and natural images, without incorporating natural videos in the training. We propose a progression of video datasets synthesized by simple generative processes, that model a growing set of natural video properties (e.g., motion, acceleration, and shape transformations). The downstream performance of video models pre-trained on these generated datasets gradually increases with the dataset progression. A VideoMAE model pre-trained on our synthetic videos closes 97.2\% of the performance gap on UCF101 action classification between training from scratch and self-supervised pre-training from natural videos, and outperforms the pre-trained model on HMDB51. Introducing crops of static images to the pre-training stage results in similar performance to UCF101 pre-training and outperforms the UCF101 pre-trained model on 11 out of 14 out-of-distribution datasets of UCF101-P. Analyzing the low-level properties of the datasets, we identify correlations between frame diversity, frame similarity to natural data, and downstream performance. Our approach provides a more controllable and transparent alternative to video data curation processes for pre-training\footnote{Project page, code and data: \url{https://unicorn53547.github.io/video_syn_rep/}}.

\end{abstract}    
\section{Introduction}
\label{sec:introduction}
Large-scale data is a fundamental component for training neural networks in various domains, such as natural language processing (NLP). To learn from such data, a prevalent technique is to pre-train models via a self-supervised task (e.g. masked modeling~\citep{bert} or next-token prediction~\citep{radford2018improving,NEURIPS2020_1457c0d6}. Adapting these models to downstream tasks results usually in improvements in various NLP tasks.

\begin{figure*}[!t]
    \centering
    \includegraphics[width=0.93\textwidth]{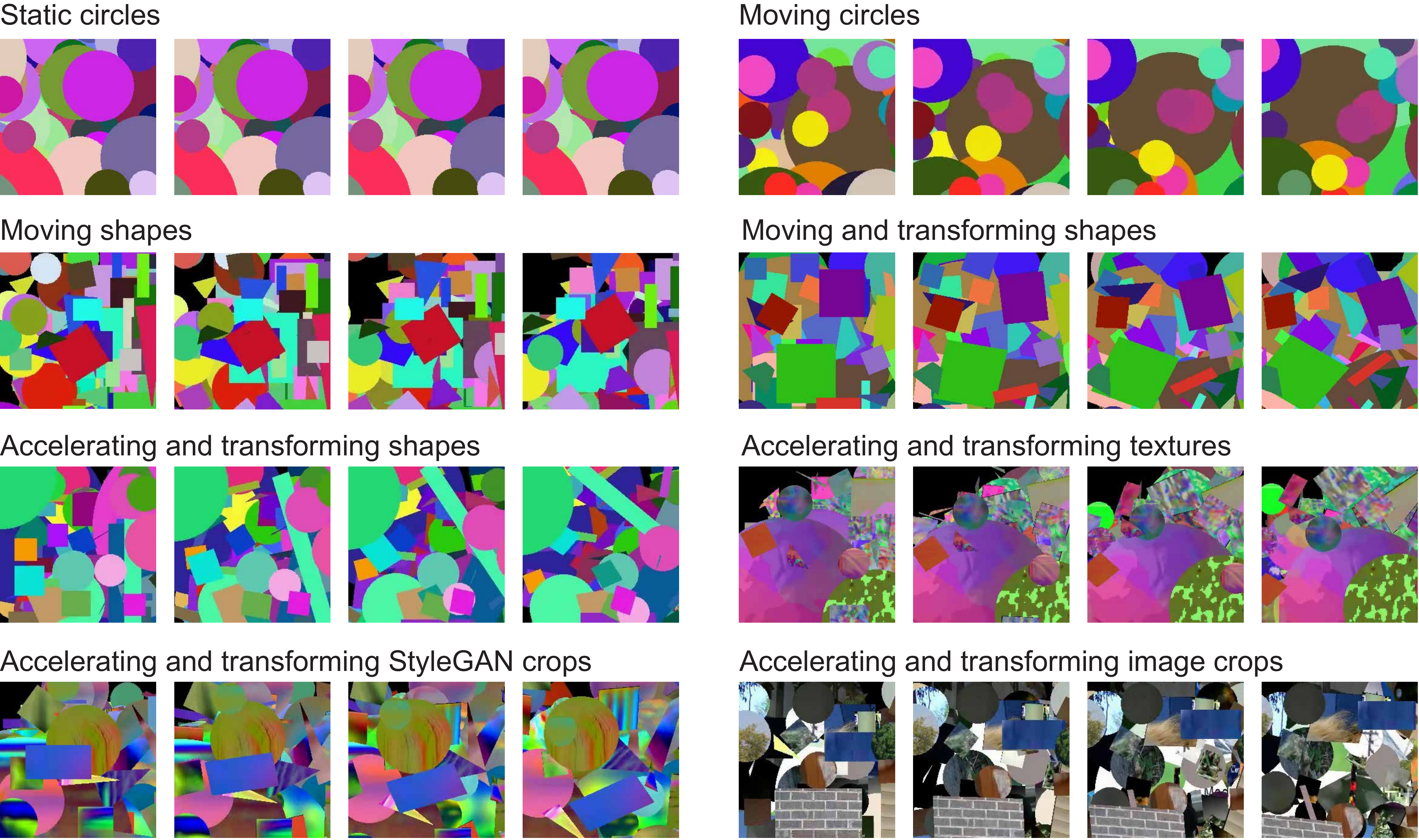}
    \vspace{-0.7em}
    \caption{\textbf{Samples from our progression of video generation models and additionally included image datasets.} We present 4 frames from timestamps $t \in \{0,10,20,30\}$ of a randomly sampled video from each of our generated datasets, and UCF101 (left to right). 
    }
    \label{fig:dataset}
\end{figure*}

While self-supervised pre-training is successful in NLP, the same level of success has not been achieved yet in computer vision. Specifically, in the video domain, although various large-scale datasets exist and have been incorporated via self-supervised learning, there is still significant room for improvements in downstream performance on video understanding (e.g., action recognition).  

One hypothesis for the limited success of self-supervised learning from videos is that current methods fail to effectively utilize all the natural video data and learn useful video representations from it. To investigate this hypothesis, we ask if natural videos are \textit{even needed} to learn video representations that are similar in performance to current state-of-the-art representations.

In this paper, we aim to reach the downstream performance of models pre-trained on natural videos, \textit{by pre-training solely on simple synthetic videos and static images}. We propose a progression of simple synthetic video generators that model a gradually growing set of video data properties -- starting from static frames with solid-color circles and introducing additional shapes, dynamics, temporal shape changes, acceleration, and other textures. We show that adding each of the different properties improves the downstream video understanding performance.

Surprisingly, we find that the gap between the performance of our models and models that were pre-trained on natural videos is minor when we pre-train using purely synthetic data and eliminated when we introduce natural
image crops. By pre-training a VideoMAE~\citep{wang2023videomaev2} on purely generated data we close 97.2\% of the gap in UCF101~\cite{soomro2012ucf101} classification accuracy between a model that was trained from scratch and a model that was pre-trained on UCF101. 
By incorporating additional crops from \textit{static images}, the performance of our models matches or improves upon the performance of the UCF101 pre-trained model, and closes 86.5\% of the gap in Kinetics-400~\cite{kay2017kinetics} between supervised training and self-supervised pre-training on Kinetics-400.

When evaluating performance on an out-of-distribution dataset, UCF101-P~\citep{robustness2022large}, \textit{the last model in our progression outperforms a model that was pre-trained on UCF101 on 11 out of 14 corrupted dataset versions}. This shows the additional benefit of training on synthetic data, and that representations of current state-of-the-art models are less reliable in out-of-distribution settings than our alternative approach, for which the generation process is fully transparent.

Finally, by comparing the accuracy of models pre-trained on the generated data in the progression, we identify different data properties that correspond with improved downstream performance. We find that high velocities and accelerations of moving shapes in the video, as well as similarity in the color space to natural videos and high frame diversity, correlate to better action recognition accuracy. We believe that these observations can help to guide future practices for large-scale self-supervised video learning.
\section{Related Work}
\label{sec:related_work}

\textbf{Video representation learning.} Early methods for learning useful video representations used models that were pre-trained on image datasets and fine-tuned on videos~\citep{twostream,spatiotemporal}.  Following the success of self-supervised representation learning (SSL) for images, similar approaches were applied to learn from videos. Earlier SSL approaches designed pretext tasks that rely on known video properties (e.g. temporal smoothness) - classifying videos to ordered or shuffled frames~\citep{misra2016shuffle,xu2019self}, predicting the future frames~\citep{mathieu2016deep}, predicting the arrow of time~\citep{wei2018learning}, and predicting video speed~\citep{benaim2020speednet}.

Recently, VideoMAE~\citep{tong2022videomae}, MAE-ST~\citep{MaskedAutoencodersSpatiotemporal2022}, and VideoMAE-V2~\citep{wang2023videomaev2} used variations of masked auto-encoding~\citep{mae} and trained a transformer to predict masked temporal video patches as the pretext task. These approaches were shown to produce useful representations without using augmentations during training. We pre-train VideoMAE models on our generated datasets and evaluate them on action recognition tasks.

\textbf{Learning from synthetic videos.} Synthetic video data is widely used for solving low-level video tasks. Specifically, data generated by 3D simulators (e.g. video game engines), were shown to be useful data sources for training models on optical flow~\citep{DFIB15} and point tracking~\citep{zheng2023point} as ground-truth labels for these tasks can be computed from the simulators. Guo \etal~\cite{9879578} pre-trained a contrastive model on videos generated from a game simulator to learn representations for human motion.  Kim \etal~\cite{kim2022how} explored how transferable are video representations learned from synthetic video data of public 3D assets. Differently, we use only \textit{simple} generative models that aim to mimic \textit{known properties} of natural videos to analyze what are the elements that enable useful video representation learning. 

\textbf{Analyzing dataset curation processes.} The research attention toward curating and characterizing useful pre-training datasets has grown recently.  Various approaches were proposed for summarizing the properties of such datasets. Dataset distillation approaches aim to summarize the datasets into a few examples that lead to the same model performance as the original datasets after training~\citep{cazenavette2022distillation,tongzhouw2018datasetdistillation}. Gadre \etal~\cite{gadre2023datacomp} introduced a benchmark for evaluating different dataset curation processes used for learning downstream tasks. Fang \etal~\cite{fang2023data} explored the correlation between data filtering heuristics and downstream performance for image classification, and proposed data filtering networks to improve filtering. 

Closer to our work, Baradad \etal~\cite{baradad2021learning,baradad2022procedural} proposed a progression of generative \textit{image} models for exploring the data properties that can unlock effective model pre-training. We follow a similar approach for video pre-training and propose a progression of generative \textit{video} models. However, unlike Baradad \etal~\cite{baradad2021learning}, each model in our progression is built \textit{on top} of the previous model. 
\section{Learning Video Representations without Natural Videos}
\label{sec:method}
To close the gap between training from scratch and natural video pre-training, and to find the key elements in data that enable useful video representation learning, we provide a progression of datasets. The datasets gradually introduce different aspects that appear in video data (e.g. transforming shapes, accelerating shapes). We pre-train SSL models on each of the generated datasets and evaluate them on downstream tasks. We present the progression of datasets and describe the generative processes that create them (\Cref{sec:sub:progression}). Then, we present the pre-training and downstream evaluation suit (\Cref{sec:sub:exp_setup}).

\subsection{Progression of video generation processes}
\label{sec:sub:progression}
We start by describing the progression of generative models $\{G_i\}$ we use to generate our training datasets. Each model uses a random number generator to sample latent parameters. The latent parameters are used for generating videos - sequences of $T$ frames $f_t \in \mathbb{R}^{H \times W \times 3}, t \in \{1,...,T\}$. Each consecutive model is built on top of the previous model, by modifying one aspect of it and adding additional calls to the random number generator. Examples of frames sampled from videos in the progression are shown in \Cref{fig:dataset}. The models in the progression are described next (see \Cref{appendix:data} for additional hyper-parameters, and the \href{https://unicorn53547.github.io/video_syn_rep/}{project page} for videos).

\textbf{Static circles.} Our first video model is of static synthetic images of multiple circles that are copied $T$ times (e.g. $f_t=f_{t+1}$). The frames are generated by positioning multiple overlapping circles on the frame canvas. The color and location of the circles are sampled uniformly at random. Following the Dead Leaves model~\citep{deadleaves}, the radius is sampled from an exponential distribution, as this distribution resembles the distribution of objects in natural scenes.

\textbf{Moving circles.} Starting from randomly positioned circles in the first frame, each assigned a velocity to derive the next frames by modeling the dynamics. Each circle is assigned a random direction and a velocity magnitude that is sampled uniformly from a fixed range. Each circle is assigned a random z-buffer value, according to the order in which it was positioned on the canvas for the first frame. This depth assignment results in occlusions when objects are moving. Introducing changes in the temporal dimension allows us to evaluate the importance of dynamics for video understanding tasks.

\textbf{Moving shapes.} We replace the circles sampled for the first frame with different shapes, including circles, quadrilaterals, and triangles. The shape types are sampled uniformly at random, and velocities are applied to them to simulate the next frames, similarly to the previous model.

\textbf{Moving and transforming shapes.} We introduce temporal transformations to the sampled shapes and apply them together with the velocities to derive the next frames. Each shape is assigned uniformly at random two scaling factors (one for each spatial dimension), a rotation speed, and two sheer factors. Each consecutive frame is computed by scaling the object in the current frame by the scaling factors, rotating it, and applying the shear mapping.

\textbf{Accelerating transforming shapes.} To introduce more complex dynamics, each temporally transforming shape is accelerated during the video by a random factor. The acceleration value is sampled uniformly from a fixed range that includes both positive and negative values.

\textbf{Accelerating transforming textures.} We replace the solid-colored shapes from the previous dataset with textures, to integrate realistic image patterns into videos. We utilize synthetic texture images from the statistical image dataset~\citep{baradad2021learning}. This dataset mimics color distribution, spectral components, and wavelet distribution characteristics of natural images and was shown to be useful for image pre-training. 
We use a total of 300k textures and for each of the shapes in the previous dataset in the progression, we sample a random texture to replace its solid color. 

\textbf{Accelerating transforming StyleGAN crops.} We replace the statistical textures with texture crops from the StyleGAN-Oriented dataset~\citep{baradad2021learning}. 
This dataset contains 300K texture images that were sampled from an untrained StyleGAN~\citep{Karras2019stylegan2} initialized to have the same wavelets for all output channels in the convolution layers. It was shown to be the most useful for \textit{image} model pre-training, out of all the synthetic datasets presented in \cite{baradad2021learning}.

\textbf{Accelerating transforming image crops.} We substitute the synthetic textures sampled for the previous Oriented-StyleGAN dataset with natural image crops, taken from ImageNet~\citep{5206848}. We do not parse or segment the images; instead, we sample random crops in the shapes mentioned above. 

\subsection{Pre-training protocol} 
\label{sec:sub:exp_setup}
We study the progression of generative models described above, by pre-training video models on sampled videos from each generator $G_i$ and evaluate them on downstream tasks. This results in a progression of pre-trained models $\{M_i\}$, where $i$ is the index of the dataset in the progression. Next, we describe our choice for pre-training model architecture, dataset sizes, and the baselines we compare to. 

\textbf{Pre-training model.} We use  VideoMAE~\citep{tong2022videomae} as our pre-training approach. Differently from other masked video auto-encoding approaches presented in \Cref{sec:related_work}, this method uses tube masking. It has been shown to outperform other SSL methods (e.g. contrastive learning approaches) without relying on heavy augmentations during pre-training.  We evaluate the pre-trained encoder of the model by fine-tuning and linear-probing it on downstream tasks. We use different model sizes to verify the consistency of the improvements in performance across scales. 

\textbf{Baselines.} We compare the pre-trained models to two additional models - a VideoMAE model that was pre-trained with the self-supervised reconstruction objective on the training data of the downstream evaluation data (UCF101/Kinetics-400), and a VideoMAE model that was initialized with random weights (e.g. trained from scratch). The former can be viewed as an upper bound for our progression, as this model is pre-trained on natural videos from the same distribution as the test set. The latter can be viewed as a lower bound, as no pre-training is done in this baseline.

\textbf{Dataset sizes and pre-training hyper-parameters.} We use the same hyperparameters as in the original pre-training recipe. That includes the same number of training steps and fine-tuning/linear probing steps. While we can generate infinite datasets from the generative models we described above, we aim to be comparable to the original pre-training dataset (UCF101). Therefore, for all the generative models that use textures or image crops, we generate sets with a similar size to the original pre-training dataset. For the other datasets, as the model manages to memorize the training data if the size is similar to the pre-training dataset, we generate random examples on the fly.

\subsection{Evaluation protocols}
We evaluate our pre-trained models for action recognition. We test the models on UCF101~\citep{soomro2012ucf101}, %a dataset that contains 13,320 video clips of human actions categorized to 101 classes, on 
HMDB51~\citep{Kuehne11}, %a dataset of additional 6,766 human action video clips categorized to 51 classes 
and Kinetics-400~\citep{kay2017kinetics}.
%with 400 human action classes, with at least 400 video clips for each action
We evaluate out-of-distribution action recognition on UCF101-P~\citep{robustness2022large}, which includes videos from the test-set of UCF101, corrupted with 4 types of low-level synthetic corruptions - camera motion, blur, noise, and digital corruptions.

As mentioned in \Cref{sec:sub:exp_setup}, each model is pre-trained and fine-tuned for the same number of steps and with the same hyper-parameters (provided in \Cref{appendix:config}). The length of the videos and the width and height are sampled to be similar to UCF101. We use the official UCF101 pre-trained checkpoint of VideoMAE as our baseline for UCF101 and HMDB51, and the official Kinetics-400 pre-trained checkpoint for Kinetics-400. 
\section{Experimental Results}

We analyze how pre-training on data sampled from the generative models presented in \cref{sec:method} affects the downstream performance. We show results for fine-tuned models on in-distribution and out-of-distribution datasets~(\Cref{sec:sub:finetune,sec:sub:ood}) and for linear-probed models (\Cref{sec:sub:linprob}).

\label{sec:experiments}
\begin{figure*}[!t]
\vspace{-1em}
    \centering
    \includegraphics[width=1.0\textwidth]{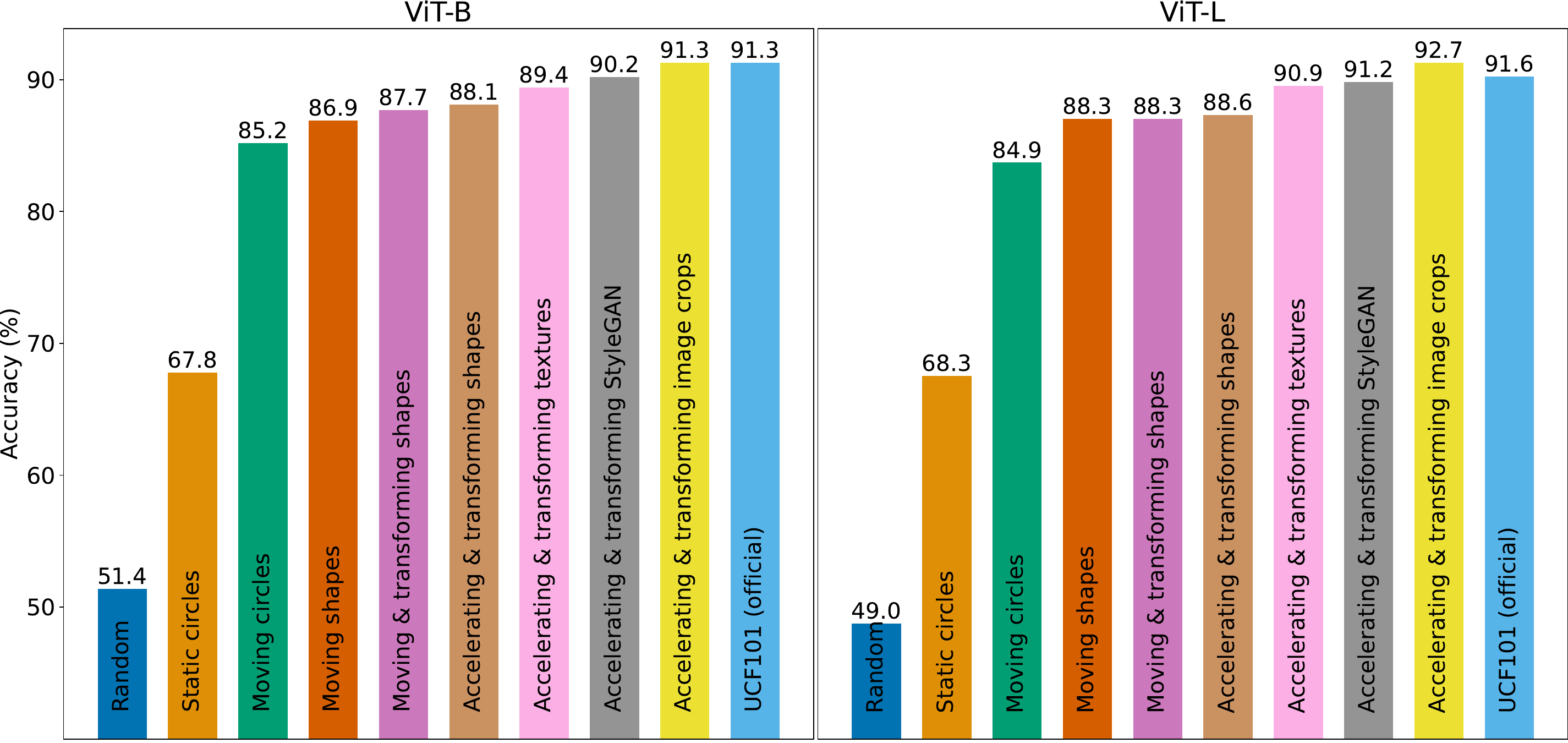}
    \vspace{-1.5em}
    \caption{\textbf{Action recognition accuracy on UCF101.} We present the UCF101 classification accuracy of the progression of models $\{M_i\}$, after fine-tuning each of them on UCF101. The accuracy increases along the progression.}
    \label{fig:results_vit_all}
\end{figure*}

\subsection{Fine-tuning} 
\label{sec:sub:finetune}
We fine-tune the pre-trained models for two different model scales, ViT-B and ViT-L, and evaluate the action recognition accuracy on UCF101, HMDB51, and Kinetics-400. We follow the protocol and hyper-parameters of \cite{tong2022videomae} and tune only the learning rate and batch size.

\textbf{UCF101 action classification.} The results are presented in \Cref{fig:results_vit_all}. The final model in the progression, accelerating and transforming shapes with ImageNet crops, performs similarly to the model that was pre-trained on the UCF101 dataset (ViT-B), or outperforms it (ViT-L). Each fine-tuned model $M_i$ in the progression improves over its predecessor, for both model scales. A large increase in performance happens when dynamics are introduced to the generated data (e.g. from static circles to moving circles). 

\textbf{HMDB51 action classification.} We evaluate the pre-trained models by fine-tuning them on the HMDB51 and present the results for ViT-B in \Cref{tab:linprob}. As shown, the order of the progression for the classification accuracy is similar. The two last models in our progression are more accurate than the model that was pre-training on UCF101. 

\textbf{Kinetics-400 action classification.} Due to the size of the dataset and our computing limitations, we only evaluated the last model in our progression and compared it to a model trained from scratch on Kinetics-400 classification and to the official VideoMAE pre-trained model on Kinetics-400. As shown in ~\Cref{tab:k400}, a model that was trained on our dataset (accelerating and transforming crops, taken from 1.3M ImageNet images) achieves an accuracy of 79.1\%, closing 86.5\% of the gap between supervised training (68.8\%) and self-supervised pre-training (80.7\%). This demonstrates that training on our datasets can achieve competitive results even when compared to a larger dataset (Kinetics-400 is 20 times larger than UCF101).

\textbf{Comparison to synthetic image pre-training for image classification.} The \textit{image} model from \citealt{baradad2021learning} that achieved the best performance after pre-training on synthetic data and fine-tuning on ImageNet classification task~\citep{5206848}, has an accuracy of 74.0\%. Compared to the baseline model that was randomly initialized (with an accuracy of 60.5\% after fine-tuning), and to a model that was pre-trained on real ImageNet data (with an accuracy of 76.1\%), the model trained on the synthetic data closes 86.5\% of the gap. For UCF101, our ViT-B model closes 97.2\% of the gap when using crops from the StyleGAN synthetic dataset, and reaches the same accuracy as the UCF101 pre-trained model with image crops (with the same size datasets as the pre-training data). This suggests that, unlike image SSL models, current video SSL models do not utilize the natural data efficiently and that most of the performance can be recovered by training on synthetic data coming from simple generative processes. 

\subsection{Distribution shift}
\label{sec:sub:ood}

\begin{figure}[!t]
    \centering
    \includegraphics[width=1.0\linewidth]{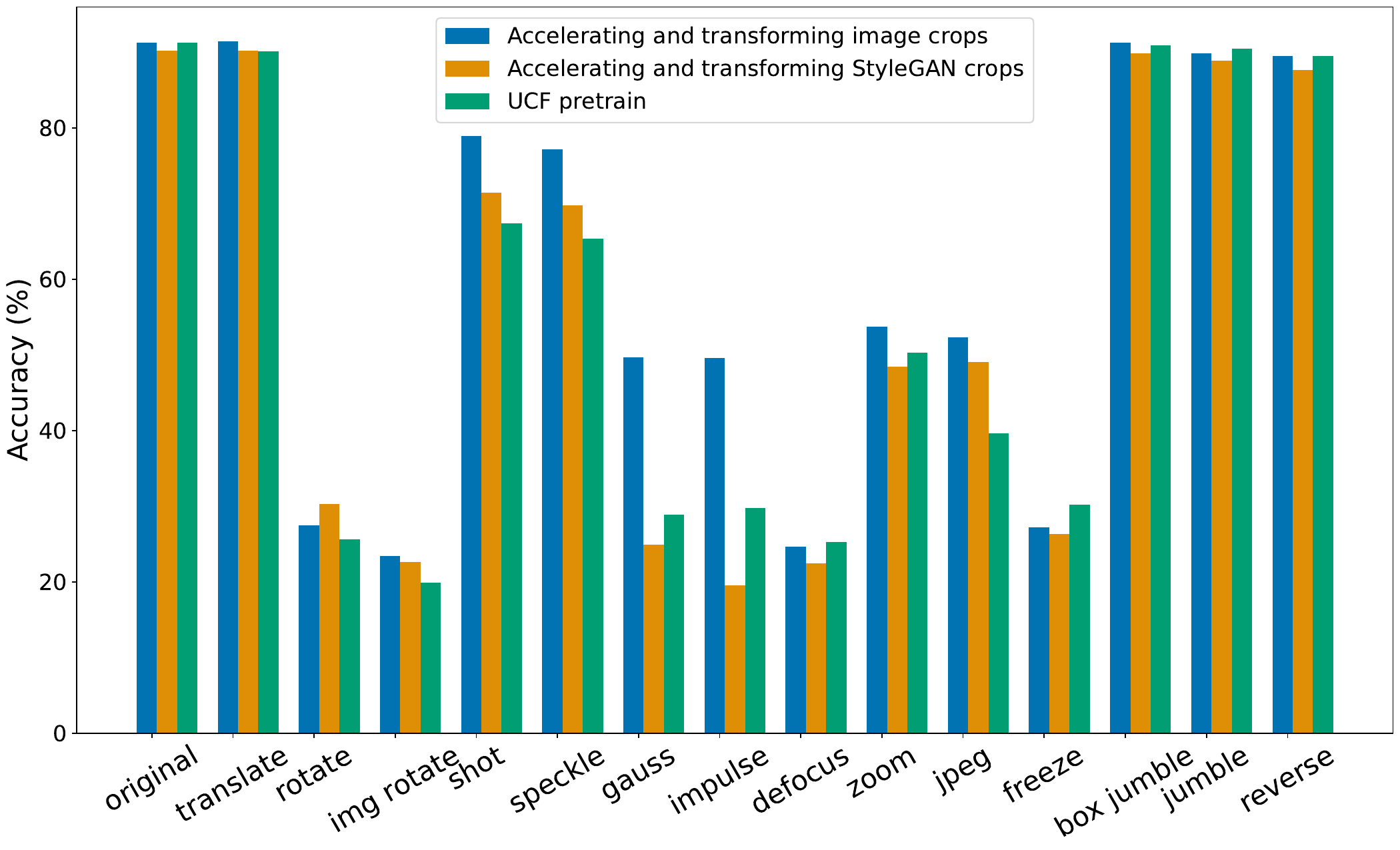}
    \vspace{-1em}
    \caption{\textbf{Distribution Shift results on UCF101-P~\citep{robustness2022large} (ViT-B)} The last model in our progression outperforms pre-training on natural videos for 11 out of 14 corruption datasets.}
    \label{fig:ood}
\end{figure}

We fine-tune the pre-trained models $\{M_i\}$ on UCF-101, and evaluate on corrupted datasets from UCF101-P~\citep{robustness2022large}. The results for the last two models in the progression are presented in \Cref{fig:ood}.
As shown, the last model in the progression outperforms the UCF101 pre-trained model on 11 out of 14 tasks and performs comparably on the rest. This suggests that the current pre-train recipe fails to generalize to out-of-distribution datasets. We note that the second to last model in our progression, which does not use real images, performs better only on 6 out of the 14 datasets. This suggests that differently from StyleGAN textures, the natural image crops unlock generalization capabilities to out-of-distribution \textit{video} corruptions.

\subsection{Linear-probing} 
\label{sec:sub:linprob}
We probe the progression of pre-trained models on UCF101. We use the same hyper-parameters as used for fine-tuning, replacing only the base learning rate to $0.01$, and do not use weight decay. The results are presented in \Cref{tab:linprob}. 

There are two main differences in the results after linear probing when compared to the results after fine-tuning. First, the difference in performance between the last model in the progression and the model trained on UCF101 is more significant (a gap of 23.2\%). Compared to the best model of \citep{baradad2021learning} that was trained on synthetic image data, which closes 56.5\% of the gap between linear probing on randomly initialized weights and linear probing on a pre-trained model, the last model in our progression closes only 40.6\% of the gap. We suspect that the difference in the gap between fine-tuning and linear probing is due to large differences between low-level properties of natural images and our datasets, which can be mitigated by fine-tuning the full model. We analyze these low-level properties in \Cref{sec:sub:static_prop}.

The second difference is that there is the progression order (when sorting by accuracy). Specifically, in contrast to fine-tuning (both on HMDB51 and UCF101), introducing gradual transformations to the shapes decreases the linear probing performance of the model compared to the previous dataset. Moreover, the order between the rest of the consecutive models in the progression is different, although the differences in performance are small. Finally, the model that uses synthetic StyleGAN crops performs better than the last model in the progression.

\begin{table}[!t]
\centering
\vspace{-0.5em}
\small
\resizebox{0.5\textwidth}{!}{
\begin{tabular}{l|c|c|c}
\toprule
 & HMDB51 & UCF101 & UCF101\\
& fine-tune & lin. prob & fine-tune\\
   \midrule
   Random initialization &18.2 & 8.9 & 51.4 \\
    Static circles & 29.2  &13.2 & 67.8\\
    Moving circles & 52.0 &15.5 & 85.2 \\
    Moving shapes & 56.1 &20.4 & 86.9 \\
    Moving and transforming shapes & 57.6 & 18.8 & 87.7 \\
    Acc. and transforming shapes & 58.9 & 18.9 & 88.1\\
    Acc. and transforming textures & 62.4 &20.9 & 89.4\\
    Acc. and transforming StyleGAN crops & \textbf{64.1} &\underline{25.2} & \underline{90.2} \\
    Acc. and transforming image crops & \textbf{64.1} &24.8 & \textbf{91.3} \\
    \midrule
    UCF101 & \underline{63.0} & \textbf{48.0} & \textbf{91.3} \\
    \bottomrule
\end{tabular}} 
\vspace{-0.5em}
\caption{\textbf{Additional action recognition results (ViT-B).} We present the classification accuracy on HMDB51 after fine-tuning and on UCF101 after linear probing/fine-tuning for all the pre-training datasets in our progression and the two baselines.}
\label{tab:linprob}
\end{table}
\section{Datasets Analysis}

In this section, we analyze in depth a few characteristics of the synthetic datasets that were shown to be useful for video pre-training. We start by evaluating the effect of incorporating natural images in the training. Then, we analyze the effects of different types of synthetic textures. Finally, we compare the statistical properties of videos to the downstream performance.

\subsection{Incorporating static images}
\label{sec:sub:imagenet}
Following the improvement of the model performance when natural image crops are used in the pre-training data, we ask: 1) how does the size of the static image dataset affect the downstream performance, 2) can the pre-training benefit from both synthetic and natural texture crops, and 3) are there alternative ways to incorporate natural images in the pre-training regime? Next, we address these questions.

\textbf{Image dataset size.} We evaluate the effect of the image data size on the downstream task. Our initial pool of images includes all the images from ImageNet (1.3M). We provide additional results with a pool with 300k images \textit{while keeping the size of the pre-training video dataset fixed}. We use the same acceleration, speed, and shape transformations as in the last dataset in the progression.
 The results for ViT-B, fine-tuned for the UCF101 classification task, are presented in \Cref{tab:imagenet_analysis}. An increase in the static images dataset results in a better performance on the downstream task.

\textbf{Combining natural images and synthetic textures.} To evaluate if useful pre-training can be achieved by combining natural images and synthetic textures, we create a dataset that incorporates crops from half of the images and crops from half of the synthetic textures from the StyleGAN textures~\citep{baradad2021learning} that we used in the previous dataset in the progression (each has 150k examples). We apply the same acceleration, speed, and transformations as in the last dataset in the progression. As shown in ~\Cref{tab:imagenet_analysis}, the performance of the new dataset (``150k images \& 150k StyleGAN'') is slightly higher than the performance of the two datasets that use solely one type of data. This suggests that mixing datasets can lead to improved performance in other cases as well. We leave this approach to future work.

\textbf{Mixing static videos of repeating single images.} We present an alternative approach to incorporate natural images into the dataset - instead of cropping images, we use full images and create videos from them by repeating the same image across all the frames. We append these static videos to the ``Accelerating and transforming shapes'' dataset, to make them the only source of textures. Their ratio in the mixed dataset is 5\% (as we found this ratio to be optimal for downstream tasks).

While the downstream model performs better than the model that was trained on the ``Accelerating and transforming shapes'' dataset (see ``Replacing 5\% of videos w/ static images'' in \Cref{tab:imagenet_analysis}), the model performs worse than using texture crops or image crops.

\begin{table}[t]
\vspace{-0.5em}
\small 
    \centering
  \begin{tabular}{l|c}
    \toprule
    \textbf{Configuration} & \textbf{Accuracy (\%)}\\
    \midrule
    300k images & 90.5\\
    150k images \& 150k StyleGAN & 90.6\\
    300k StyleGAN & 90.2 \\
    300k statistical textures & 89.4 \\
    \midrule
    1.3M images & 91.3 \\ 
   \midrule
    Replacing 5\% of videos  & \multirow{ 2}{*} {{88.5}} \\
    w/ static images & \\
    \bottomrule
  \end{tabular}
  \caption{\textbf{Incorporating natural images into training (ViT-B).} We ablate different approaches for incorporating natural images during training, and evaluate them on UCF101. }
  \label{tab:imagenet_analysis}
\end{table}
\begin{table}
\centering
  \small
  \begin{tabular}{l|c}
    \toprule
    \textbf{Configuration  (ViT-B)} & \textbf{Accuracy (\%)}\\
    \midrule
    Static StyleGAN crops & 90.2 \\ 
    Dynamic StyleGAN crops & 89.2\\
    Dynamic StyleGAN videos & {68.7} \\
    \bottomrule
  \end{tabular}
  \vspace{-0.5em}
  \caption{\textbf{Incorporating synthetic textures into training.} Introducing dynamics to the StyleGAN textures does not improve performance.\label{tab:textures_analysis}}

\vspace{-1em}
\end{table}

\subsection{Incorporating dynamic textures}
\label{sec:sub:textures}

We evaluate additional synthetic dynamics that can be incorporated in our progression. Specifically, we replace the static StyleGAN textures with a dynamic version.  

\textbf{Dynamic StyleGAN textures.} We investigate a simple extension of the StyleGAN-generated textures into videos. We create a texture video by starting from a random noise $z_0$, provided as a latent code to the StyleGAN generator $G'$. Each consecutive frame is generated by adding a random noise with a smaller standard deviation $\delta z_{i}$ to the previous latent $z_{i-1}$  ($z_{i} = z_{i-1} +\delta z_{i}$) and generating a frame $G'(z_i)$. We explore two approaches to incorporate these texture videos: directly creating a dataset with multiple such videos (``Dynamic StyleGAN videos'') or replacing the solid-color shapes from the ``accelerating and transforming shapes'' with dynamic texture crops that are updated across frames (``Dynamic StyleGAN crops''). 

\textbf{Fine-tuning on UCF101.} \Cref{tab:textures_analysis} presents the action classification accuracy after incorporating the dynamic textures in the pre-training stage and fine-tuning on UCF101. Using Dynamic StyleGAN videos leads to performance that is only slightly better than training on static circles (67.8\%). Replacing the static StyleGAN crops with \textit{dynamic} StyleGAN crops leads to a performance drop of 1\%. Both results suggest that the simple hand-crafted dynamics of randomly moving Dead-Leaves models are sufficient for pre-training, without introducing additional dynamics modeling.

\subsection{Similarity to Pre-training Dataset}

\label{sec:sub:dynamic} 
During our experiments, we created multiple versions of each dataset we presented, with differences in configuration (e.g. different video background colors and different object speeds). In total, we generated 28 datasets and trained ViT-B VideoMAE on each (see \Cref{app:additional_datasets}). 
We plot the UCF101 fine-tuning accuracies of the models as a function of their similarity to UCF101. We compute single-frame similarity and video similarity.

\textbf{FID.} We compare the similarity between \textit{frames} from our datasets and UCF101 to the classification accuracy. We compute FID~\citep{NIPS2017_8a1d6947} on randomly sampled frames (10k frames from each dataset). FID is a common metric for evaluating the similarity between two image datasets by comparing the Frechet distance between distributions of deep features extracted from them. 

As shown in~\Cref{fig:all_property}.a there is a strong negative correlation between the frame similarity to the accuracy ($r=-0.72$). This suggests that improving frame similarity can lead to better performance. Nevertheless, the FID scores are considered to be high, and our datasets are significantly different from the original UCF101 data.  

\textbf{FVD.} In this analysis we compare the classification accuracy to the \textit{video} similarity between our datasets and UCF101. We compute FVD~\citep{Unterthiner2019FVDAN} on 1,000 random videos from each of the datasets and present the results in~\Cref{fig:all_property}.b. Differently from the frame similarity, there is less significant negative correlation between the FVD metric and the performance ($r=-0.27$). This suggests that this metric is less indicative of downstream performance. 

\subsection{Static properties of individual frames}
\label{sec:sub:static_prop}
We follow \citep{baradad2021learning} and compare the properties of individual frames in the 28 datasets that we generated to the downstream performance. Similarly to \Cref{sec:sub:dynamic}, we randomly sample 1000 videos from all the datasets we analyzed and compare low-level statistics to the downstream classification accuracy. 

\begin{figure*}
    \centering
    \includegraphics[width=0.99\textwidth]{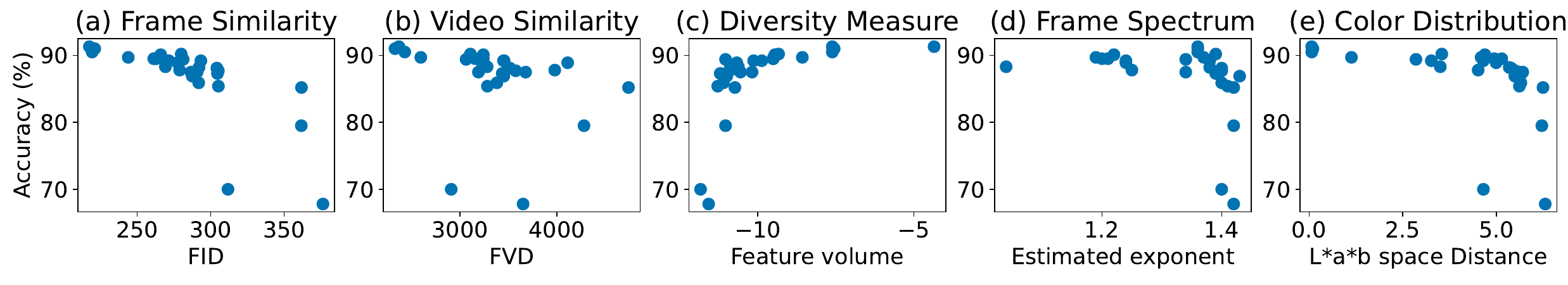}
    \vspace{-1.em}
    \caption{\textbf{Dataset properties compared to downstream performance.} We compare the downstream classification accuracy on UCF101 after fine-tuning to frame and video properties of all the dataset variants we used in our analysis (see datasets list in~\Cref{appendix:data}).
    % \xueyang{$r=-0.72, -0.27, 0.53, -0.35, -0.42$ from left to right}
    \label{fig:all_property}}
\end{figure*}

\textbf{Diversity.} We measure the diversity of the frames in the dataset. We utilize inception features~\citep{DBLP:journals/corr/SzegedyVISW15} computed for 16 sampled frames in randomly sampled videos and plot the determinant of their covariance matrix. As shown in \Cref{fig:all_property}.c, There is a moderate correlation between the accuracy and the diversity ($r=0.53$). According to this measure, all the generated datasets are less diverse than UCF101. Nevertheless, the datasets that include synthetic textures (statistical or StyleGAN-based) and the ones that include image crops are more diverse than the other datasets. This suggests that investing in more diverse datasets can improve performance even further.

\textbf{Image spectrum.}  Following \cite{statsofimages}, that showed that the spectrum of natural images resembles the function $A/{|f|}^{\alpha}$, with a scaling factor $A$ and an exponent $\alpha$ ranging in $[0.5, 2.0]$, we estimate the exponent for frames in our datasets. The results are presented in \Cref{fig:all_property}.d. The datasets that result in the best downstream performance have an estimated exponent that lies close to the middle of the range, between 1.2 and 1.4.

\textbf{Color statistics.} We compare the distance in color space between the generated data and natural videos. We compute the symmetric KL divergence between the color distributions of each dataset. We model the color distributions as three-dimensional Gaussian that correspond to the three color channels in L*a*b space. \Cref{fig:all_property}.e presents the distances between UCF101 color statistics and the datasets in our progression. There is a relatively weak negative correlation of $r=-0.42$ between the color distance to UCF101 and the accuracy.

\begin{figure*}[!t]
    \centering
    \includegraphics[width=0.95\linewidth]{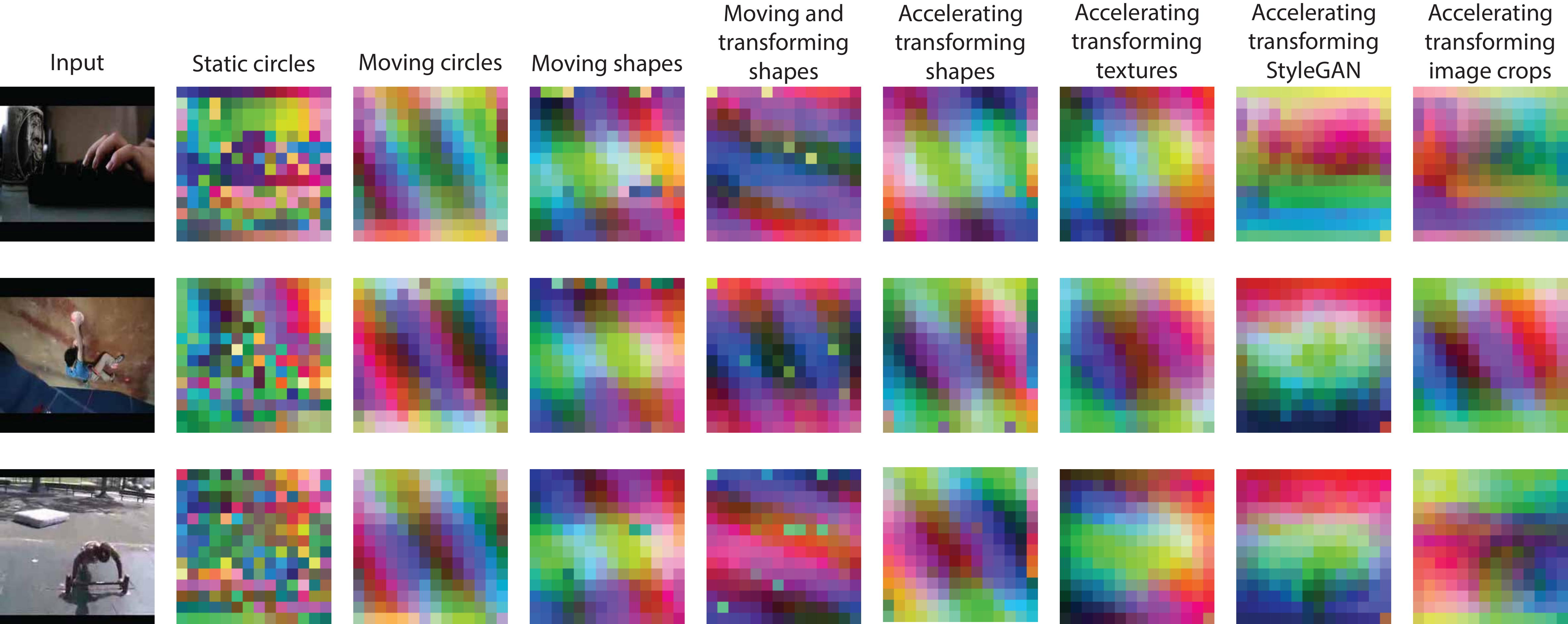}
    \vspace{-.5em}
    \caption{\textbf{Feature visualizations for pre-trained models.} We present the 3 principal components of the attention keys of the last encoder layer, for all $M_i$ as the three color channels. Different object parts start to appear as the datasets progress.}
    \label{fig:pca}
\end{figure*}
\subsection{Representation visualization}
We visualize the learned representation produced by the models $M_i$. Following \cite{amir2021deep}, we compute PCA on the attention keys extracted from the last VideoMAE encoder layer across 32 frames from 70 videos from the same class of UCF101. We plot the first three principal components as red, green, and blue channels and present features for 2-frame inputs (see full PCA videos in the \href{https://unicorn53547.github.io/video_syn_rep/}{project page}). 

The visualizations for videos from three classes are presented in \Cref{fig:pca}. The PCs of the features produced by the pre-trained models are relatively different. While the early models in the progression capture mostly static positional information about the frames, later models preserve some structural information in the input in the 3 PCs. 

\section{Limitations}

\textbf{Generalization to other tasks.} While the pre-trained models are evaluated on different datasets (HMDB51, UCF101, Kintetics, and UCF101-P) and with two different adaptation regimes (linear probing and fine-tuning), and show relatively similar trends across the progression of the datasets, they can have different trends when adapted to other tasks. We decided to focus on action recognition and to aim to reach the performance of relatively \textit{small datasets}, as a first step to a fully synthetic approach that does not rely on natural videos. We do not tune any hyper-parameters (except batch size and learning rate, due to GPU memory capacity differences) to improve performance. In future work, we aim to explore other tasks and apply training regimes that were shown to work on larger datasets.

\textbf{Generalization to other model types.} Our evaluating suite included pre-training of one type of model - VideoMAE. While this pre-training approach is widely used, the behavior we presented for different datasets may be different for other pre-training regimes. Our decision to focus on one model follows a similar scope of \cite{baradad2021learning}.

\section{Discussion} 
Learning from data produced by simple generative processes and other well-studied data sources has an advantage over learning from large-scale video data - when pre-training on large-scale video corpus, commonly obtained from the internet, it is merely impossible to monitor what are all the training examples and to verify that no malicious, private, or biased data is included in the pre-training stage. Learning from generated data, on the other hand, gives better control over the content. 

We believe that the synthetic data analysis we provided can be utilized to create better datasets for learning video representations without natural videos. Guided by the analysis, we plan to explore other well-understood data sources and generation processes to continue improving video representation learning, in large-scale training regimes.

While it was not our main goal in this paper, the synthetic data we produced can be incorporated as augmentations as well (see \Cref{tab:additional_mix_real}). Pre-training both on UCF101 \textit{and} the last dataset in the progression leads to an accuracy of 92.0\% after fine-tuning, surpassing the performance of solely pre-training on UCF101. We plan to explore this direction in the future as well.
\section{Acknowledgments}
The authors would like to thank Amil Dravid and Ren Wang for their valuable comments and feedback on our paper; and thank UC Berkeley for the computational support to perform data processing and experiments. YG is supported by the Google Fellowship. 
{
    \small
    \bibliographystyle{ieeenat_fullname}
    \bibliography{main}

\begin{thebibliography}{36}
\providecommand{\natexlab}[1]{#1}
\providecommand{\url}[1]{\texttt{#1}}
\expandafter\ifx\csname urlstyle\endcsname\relax
  \providecommand{\doi}[1]{doi: #1}\else
  \providecommand{\doi}{doi: \begingroup \urlstyle{rm}\Url}\fi

\bibitem[Amir et~al.(2023)Amir, Gandelsman, Bagon, and Dekel]{amir2021deep}
Shir Amir, Yossi Gandelsman, Shai Bagon, and Tali Dekel.
\newblock On the effectiveness of vit features as local semantic descriptors.
\newblock In \emph{Computer Vision – ECCV 2022 Workshops: Tel Aviv, Israel, October 23–27, 2022, Proceedings, Part IV}, page 39–55, Berlin, Heidelberg, 2023. Springer-Verlag.

\bibitem[Baradad et~al.(2021)Baradad, Wulff, Wang, Isola, and Torralba]{baradad2021learning}
Manel Baradad, Jonas Wulff, Tongzhou Wang, Phillip Isola, and Antonio Torralba.
\newblock Learning to see by looking at noise.
\newblock In \emph{Advances in Neural Information Processing Systems}, 2021.

\bibitem[Baradad et~al.(2022)Baradad, Chen, Wulff, Wang, Feris, Torralba, and Isola]{baradad2022procedural}
Manel Baradad, Chun-Fu Chen, Jonas Wulff, Tongzhou Wang, Rogerio Feris, Antonio Torralba, and Phillip Isola.
\newblock Procedural image programs for representation learning.
\newblock In \emph{Advances in Neural Information Processing Systems}, 2022.

\bibitem[Benaim et~al.(2020)Benaim, Ephrat, Lang, Mosseri, Freeman, Rubinstein, Irani, and Dekel]{benaim2020speednet}
Sagie Benaim, Ariel Ephrat, Oran Lang, Inbar Mosseri, William~T Freeman, Michael Rubinstein, Michal Irani, and Tali Dekel.
\newblock Speednet: Learning the speediness in videos.
\newblock In \emph{Proceedings of the IEEE/CVF conference on computer vision and pattern recognition}, pages 9922--9931, 2020.

\bibitem[Bordenave et~al.(2006)Bordenave, Gousseau, and Roueff]{deadleaves}
Charles Bordenave, Yann Gousseau, and François Roueff.
\newblock The dead leaves model: A general tessellation modeling occlusion.
\newblock \emph{Advances in Applied Probability}, 38\penalty0 (1):\penalty0 31--46, 2006.

\bibitem[Brown et~al.(2020)Brown, Mann, Ryder, Subbiah, Kaplan, Dhariwal, Neelakantan, Shyam, Sastry, Askell, Agarwal, Herbert-Voss, Krueger, Henighan, Child, Ramesh, Ziegler, Wu, Winter, Hesse, Chen, Sigler, Litwin, Gray, Chess, Clark, Berner, McCandlish, Radford, Sutskever, and Amodei]{NEURIPS2020_1457c0d6}
Tom Brown, Benjamin Mann, Nick Ryder, Melanie Subbiah, Jared~D Kaplan, Prafulla Dhariwal, Arvind Neelakantan, Pranav Shyam, Girish Sastry, Amanda Askell, Sandhini Agarwal, Ariel Herbert-Voss, Gretchen Krueger, Tom Henighan, Rewon Child, Aditya Ramesh, Daniel Ziegler, Jeffrey Wu, Clemens Winter, Chris Hesse, Mark Chen, Eric Sigler, Mateusz Litwin, Scott Gray, Benjamin Chess, Jack Clark, Christopher Berner, Sam McCandlish, Alec Radford, Ilya Sutskever, and Dario Amodei.
\newblock Language models are few-shot learners.
\newblock In \emph{Advances in Neural Information Processing Systems}, pages 1877--1901. Curran Associates, Inc., 2020.

\bibitem[Cazenavette et~al.(2022)Cazenavette, Wang, Torralba, Efros, and Zhu]{cazenavette2022distillation}
George Cazenavette, Tongzhou Wang, Antonio Torralba, Alexei~A. Efros, and Jun-Yan Zhu.
\newblock Dataset distillation by matching training trajectories.
\newblock In \emph{Proceedings of the IEEE/CVF Conference on Computer Vision and Pattern Recognition}, 2022.

\bibitem[Deng et~al.(2009)Deng, Dong, Socher, Li, Li, and Fei-Fei]{5206848}
Jia Deng, Wei Dong, Richard Socher, Li-Jia Li, Kai Li, and Li Fei-Fei.
\newblock Imagenet: A large-scale hierarchical image database.
\newblock In \emph{2009 IEEE Conference on Computer Vision and Pattern Recognition}, pages 248--255, 2009.

\bibitem[Devlin et~al.(2019)Devlin, Chang, Lee, and Toutanova]{bert}
Jacob Devlin, Ming-Wei Chang, Kenton Lee, and Kristina Toutanova.
\newblock {BERT}: Pre-training of deep bidirectional transformers for language understanding.
\newblock In \emph{Proceedings of the 2019 Conference of the North {A}merican Chapter of the Association for Computational Linguistics: Human Language Technologies, Volume 1 (Long and Short Papers)}, pages 4171--4186, Minneapolis, Minnesota, 2019. Association for Computational Linguistics.

\bibitem[Dosovitskiy et~al.(2015)Dosovitskiy, Fischer, Ilg, H{\"a}usser, Haz{\i}rba{\c{s}}, Golkov, v.d. Smagt, Cremers, and Brox]{DFIB15}
A. Dosovitskiy, P. Fischer, E. Ilg, P. H{\"a}usser, C. Haz{\i}rba{\c{s}}, V. Golkov, P. v.d. Smagt, D. Cremers, and T. Brox.
\newblock Flownet: Learning optical flow with convolutional networks.
\newblock In \emph{IEEE International Conference on Computer Vision (ICCV)}, 2015.

\bibitem[Fang et~al.(2024)Fang, Jose, Jain, Schmidt, Toshev, and Shankar]{fang2023data}
Alex Fang, Albin~Madappally Jose, Amit Jain, Ludwig Schmidt, Alexander~T Toshev, and Vaishaal Shankar.
\newblock Data filtering networks.
\newblock In \emph{The Twelfth International Conference on Learning Representations}, 2024.

\bibitem[Feichtenhofer et~al.(2022)Feichtenhofer, Li, He, et~al.]{MaskedAutoencodersSpatiotemporal2022}
Christoph Feichtenhofer, Yanghao Li, Kaiming He, et~al.
\newblock Masked autoencoders as spatiotemporal learners.
\newblock \emph{Advances in neural information processing systems}, 35:\penalty0 35946--35958, 2022.

\bibitem[Gadre et~al.(2024)Gadre, Ilharco, Fang, Hayase, Smyrnis, Nguyen, Marten, Wortsman, Ghosh, Zhang, et~al.]{gadre2023datacomp}
Samir~Yitzhak Gadre, Gabriel Ilharco, Alex Fang, Jonathan Hayase, Georgios Smyrnis, Thao Nguyen, Ryan Marten, Mitchell Wortsman, Dhruba Ghosh, Jieyu Zhang, et~al.
\newblock Datacomp: In search of the next generation of multimodal datasets.
\newblock \emph{Advances in Neural Information Processing Systems}, 36, 2024.

\bibitem[Guo et~al.(2022)Guo, Wu, Wang, Su, Su, Gan, Huang, and Yang]{9879578}
Xi Guo, Wei Wu, Dongliang Wang, Jing Su, Haisheng Su, Weihao Gan, Jian Huang, and Qin Yang.
\newblock Learning video representations of human motion from synthetic data.
\newblock In \emph{2022 IEEE/CVF Conference on Computer Vision and Pattern Recognition (CVPR)}, pages 20165--20175, 2022.

\bibitem[He et~al.(2022)He, Chen, Xie, Li, Dollar, and Girshick]{mae}
Kaiming He, Xinlei Chen, Saining Xie, Yanghao Li, Piotr Dollar, and Ross Girshick.
\newblock Masked autoencoders are scalable vision learners.
\newblock In \emph{Proceedings - 2022 IEEE/CVF Conference on Computer Vision and Pattern Recognition, CVPR 2022}, pages 15979--15988. IEEE Computer Society, 2022.
\newblock Publisher Copyright: {\textcopyright} 2022 IEEE.; 2022 IEEE/CVF Conference on Computer Vision and Pattern Recognition, CVPR 2022 ; Conference date: 19-06-2022 Through 24-06-2022.

\bibitem[Heusel et~al.(2017)Heusel, Ramsauer, Unterthiner, Nessler, and Hochreiter]{NIPS2017_8a1d6947}
Martin Heusel, Hubert Ramsauer, Thomas Unterthiner, Bernhard Nessler, and Sepp Hochreiter.
\newblock Gans trained by a two time-scale update rule converge to a local nash equilibrium.
\newblock In \emph{Advances in Neural Information Processing Systems}. Curran Associates, Inc., 2017.

\bibitem[Karras et~al.(2020)Karras, Laine, Aittala, Hellsten, Lehtinen, and Aila]{Karras2019stylegan2}
Tero Karras, Samuli Laine, Miika Aittala, Janne Hellsten, Jaakko Lehtinen, and Timo Aila.
\newblock Analyzing and improving the image quality of {StyleGAN}.
\newblock In \emph{Proc. CVPR}, 2020.

\bibitem[Kay et~al.(2017)Kay, Carreira, Simonyan, Zhang, Hillier, Vijayanarasimhan, Viola, Green, Back, Natsev, et~al.]{kay2017kinetics}
Will Kay, Joao Carreira, Karen Simonyan, Brian Zhang, Chloe Hillier, Sudheendra Vijayanarasimhan, Fabio Viola, Tim Green, Trevor Back, Paul Natsev, et~al.
\newblock The kinetics human action video dataset.
\newblock \emph{arXiv preprint arXiv:1705.06950}, 2017.

\bibitem[Kim et~al.(2022)Kim, Mishra, Jin, Panda, Kuehne, Karlinsky, Saligrama, Saenko, Oliva, and Feris]{kim2022how}
YoWhan Kim, Samarth Mishra, SouYoung Jin, Rameswar Panda, Hilde Kuehne, Leonid Karlinsky, Venkatesh Saligrama, Kate Saenko, Aude Oliva, and Rogerio Feris.
\newblock How transferable are video representations based on synthetic data?
\newblock In \emph{Thirty-sixth Conference on Neural Information Processing Systems Datasets and Benchmarks Track}, 2022.

\bibitem[Kuehne et~al.(2011)Kuehne, Jhuang, Garrote, Poggio, and Serre]{Kuehne11}
H. Kuehne, H. Jhuang, E. Garrote, T. Poggio, and T. Serre.
\newblock {HMDB}: a large video database for human motion recognition.
\newblock In \emph{Proceedings of the International Conference on Computer Vision (ICCV)}, 2011.

\bibitem[Mathieu et~al.(2015)Mathieu, Couprie, and LeCun]{mathieu2016deep}
Micha{\"e}l Mathieu, Camille Couprie, and Yann LeCun.
\newblock Deep multi-scale video prediction beyond mean square error.
\newblock \emph{CoRR}, abs/1511.05440, 2015.

\bibitem[Misra et~al.(2016)Misra, Zitnick, and Hebert]{misra2016shuffle}
Ishan Misra, C~Lawrence Zitnick, and Martial Hebert.
\newblock Shuffle and learn: unsupervised learning using temporal order verification.
\newblock In \emph{Computer Vision--ECCV 2016: 14th European Conference, Amsterdam, The Netherlands, October 11--14, 2016, Proceedings, Part I 14}, pages 527--544. Springer, 2016.

\bibitem[Radford et~al.(2018)Radford, Narasimhan, Salimans, and Sutskever]{radford2018improving}
Alec Radford, Karthik Narasimhan, Tim Salimans, and Ilya Sutskever.
\newblock Improving language understanding by generative pre-training.
\newblock 2018.

\bibitem[Schiappa et~al.(2023)Schiappa, Biyani, Kamtam, Vyas, Palangi, Vineet, and Rawat]{robustness2022large}
Madeline~C Schiappa, Naman Biyani, Prudvi Kamtam, Shruti Vyas, Hamid Palangi, Vibhav Vineet, and Yogesh Rawat.
\newblock Large-scale robustness analysis of video action recognition models.
\newblock In \emph{The IEEE/CVF Conference on Computer Vision and Pattern Recognition}, 2023.

\bibitem[Simonyan and Zisserman(2014)]{twostream}
Karen Simonyan and Andrew Zisserman.
\newblock Two-stream convolutional networks for action recognition in videos.
\newblock In \emph{Proceedings of the 27th International Conference on Neural Information Processing Systems - Volume 1}, page 568–576, Cambridge, MA, USA, 2014. MIT Press.

\bibitem[Soomro et~al.(2012)Soomro, Zamir, and Shah]{soomro2012ucf101}
Khurram Soomro, Amir Zamir, and Mubarak Shah.
\newblock Ucf101: A dataset of 101 human actions classes from videos in the wild.
\newblock \emph{ArXiv}, abs/1212.0402, 2012.

\bibitem[Szegedy et~al.(2015)Szegedy, Vanhoucke, Ioffe, Shlens, and Wojna]{DBLP:journals/corr/SzegedyVISW15}
Christian Szegedy, Vincent Vanhoucke, Sergey Ioffe, Jonathon Shlens, and Zbigniew Wojna.
\newblock Rethinking the inception architecture for computer vision.
\newblock \emph{CoRR}, abs/1512.00567, 2015.

\bibitem[Tong et~al.(2022)Tong, Song, Wang, and Wang]{tong2022videomae}
Zhan Tong, Yibing Song, Jue Wang, and Limin Wang.
\newblock Video{MAE}: Masked autoencoders are data-efficient learners for self-supervised video pre-training.
\newblock In \emph{Advances in Neural Information Processing Systems}, 2022.

\bibitem[Torralba and Oliva(2003)]{statsofimages}
Antonio Torralba and Aude Oliva.
\newblock Statistics of natural image categories.
\newblock \emph{Network: Computation in Neural Systems}, 14\penalty0 (3):\penalty0 391--412, 2003.
\newblock PMID: 12938764.

\bibitem[Tran et~al.(2018)Tran, Wang, Torresani, Ray, LeCun, and Paluri]{spatiotemporal}
Du Tran, Heng Wang, Lorenzo Torresani, Jamie Ray, Yann LeCun, and Manohar Paluri.
\newblock A closer look at spatiotemporal convolutions for action recognition.
\newblock In \emph{Proceedings of the IEEE conference on Computer Vision and Pattern Recognition}, pages 6450--6459, 2018.

\bibitem[Unterthiner et~al.(2019)Unterthiner, van Steenkiste, Kurach, Marinier, Michalski, and Gelly]{Unterthiner2019FVDAN}
Thomas Unterthiner, Sjoerd van Steenkiste, Karol Kurach, Rapha{\"e}l Marinier, Marcin Michalski, and Sylvain Gelly.
\newblock Fvd: A new metric for video generation.
\newblock In \emph{DGS@ICLR}, 2019.

\bibitem[Wang et~al.(2023)Wang, Huang, Zhao, Tong, He, Wang, Wang, and Qiao]{wang2023videomaev2}
Limin Wang, Bingkun Huang, Zhiyu Zhao, Zhan Tong, Yinan He, Yi Wang, Yali Wang, and Yu Qiao.
\newblock Videomae v2: Scaling video masked autoencoders with dual masking.
\newblock In \emph{Proceedings of the IEEE/CVF Conference on Computer Vision and Pattern Recognition (CVPR)}, pages 14549--14560, 2023.

\bibitem[Wang et~al.(2018)Wang, Zhu, Torralba, and Efros]{tongzhouw2018datasetdistillation}
Tongzhou Wang, Jun-Yan Zhu, Antonio Torralba, and Alexei~A Efros.
\newblock Dataset distillation.
\newblock \emph{arXiv preprint arXiv:1811.10959}, 2018.

\bibitem[Wei et~al.(2018)Wei, Lim, Zisserman, and Freeman]{wei2018learning}
Donglai Wei, Joseph~J Lim, Andrew Zisserman, and William~T Freeman.
\newblock Learning and using the arrow of time.
\newblock In \emph{Proceedings of the IEEE Conference on Computer Vision and Pattern Recognition}, pages 8052--8060, 2018.

\bibitem[Xu et~al.(2019)Xu, Xiao, Zhao, Shao, Xie, and Zhuang]{xu2019self}
Dejing Xu, Jun Xiao, Zhou Zhao, Jian Shao, Di Xie, and Yueting Zhuang.
\newblock Self-supervised spatiotemporal learning via video clip order prediction.
\newblock In \emph{Computer Vision and Pattern Recognition (CVPR)}, 2019.

\bibitem[Zheng et~al.(2023)Zheng, Harley, Shen, Wetzstein, and Guibas]{zheng2023point}
Yang Zheng, Adam~W. Harley, Bokui Shen, Gordon Wetzstein, and Leonidas~J. Guibas.
\newblock Pointodyssey: A large-scale synthetic dataset for long-term point tracking.
\newblock In \emph{ICCV}, 2023.

\end{thebibliography}
}

\clearpage
\setcounter{page}{1}
\maketitlesupplementary

\section{Appendix}
\subsection{Results on Kinetics-400}
We conduct experiments on a much larger dataset Kinetics-400~\citep{kay2017kinetics} and report the results in \Cref{tab:k400}. For this dataset, we close 86.6\% of the gap between training from scratch and self-supervised pre-training from natural videos.
\begin{table}[!htbp]
  \centering
  \begin{tabular}{l|c}
    \toprule
    Pre-training Dataset &  Accuracy \\
    \midrule
    Scratch & 68.8 \\
    Accelerating and transforming image crops & 79.1\\
    Kinetics-400 & 80.7 \\
    \bottomrule
  \end{tabular}
  \caption{\textbf{Results on Kinetics-400 test set~\citep{kay2017kinetics}}. The kinetics-400 result is obtained by fine-tuning from the official pre-trained VideoMAE checkpoint~\citep{tong2022videomae}. }
  \label{tab:k400}
\end{table}   
\subsection{Additional dataset details}

\label{appendix:data}
We provide the hyper-parameter configuration for the dataset generators in \Cref{tab:Dataset}. We provide additional explanations next. Please see the attached HTML for videos of the datasets in our progression.

\textbf{Dataset size.} For datasets without textures or image crops, we use an on-the-fly generation strategy for training. For video data with textures and image crops, we generate 9537 videos for training, the same number in the UCF101 training set. For all each generated video, we use a resolution of $256\times 256$ for, FPS of 25, and a duration that is sampled uniformly in $(100, 200)$.

\textbf{Acceleration and speed parameters.} For each object, we sample its absolute speed from a uniform distribution ranging between $1.2$ and $3,0$ pixels per time frame. The absolute acceleration sampled from a uniform distribution $(-0.6, 0.6)$. The moving direction sampled uniformly between $(-\pi, \pi)$.

\textbf{Transformation parameters.} To introduce dynamics additionally to translation, we apply scale, shear, and rotation transformations. By default, the rotation angle is set to a uniform distribution of $(-\frac{1}{100}\pi, \frac{1}{100}\pi)$, scale and shear factors are set to a randomly chosen number from $(-0.005, 0.005)$ in  both x-axis and y-axis.

\subsection{Training configuration}
We provide the hyperparameter configurations for pre-training (\Cref{tab:Pretrain_setting}), fine-tuning on UCF101 (\Cref{tab:Finetune}), and linear probing (\Cref{tab:Linear Probe}) on the ViT-B model. The configuration for fine-tuning is similar to the original configuration from \citealt{tong2022videomae}, except for the batch-size, learning rate, and the Adam optimizer hyper-parameters. The fine-tuning configuration on HMDB51 is the same as for UCF101, except for the number of test clips, which is 10. The pre-training and fine-tuning setting for ViT-L is same with ViT-B, except for reducing batch size to half.
\label{appendix:config}
\begin{table}[h]
\small
% \tablestyle{1pt}{1.02}
\begin{center}
\begin{tabular}{c|c}
\toprule
Hyperparameter & Value \\
\midrule
masking ratio & 0.75 \\
training epochs & 3200 \\
optimizer & {AdamW} \\ 
base learning & {3e-4}\\
weight decay & {0.05} \\
optimizer momentum & {$\beta_1=0.9, \beta_2=0.95$} \\
batch size & {256} \\
learning rate schedule & cosine decay \\ %~\cite{coslr} \\
warmup epochs & {40} \\
augmentation & {MultiScaleCrop}\\ %~\cite{tsn_journal}} \\
\bottomrule
\end{tabular}
\end{center}
% \vspace{.2em}
\caption{\textbf{Pre-training settings (ViT-B).}}
\label{tab:Pretrain_setting} 
\end{table}

\begin{table}[h]
\small
% \tablestyle{1pt}{1.02}
\begin{center}
\begin{tabular}{c|c}
\toprule
Hyperparameter & Value \\
\midrule
training epochs & 100 \\
optimizer & {AdamW} \\ 
base learning & {1e-3}\\
weight decay & {0.05} \\
optimizer momentum & {$\beta_1=0.9, \beta_2=0.95$} \\
batch size & {256} \\
learning rate schedule & cosine decay \\ %~\cite{coslr} \\
warmup epochs & {5} \\
flip augmentation & yes \\
RandAug  & (9, 0.5) \\
label smoothing & 0.1 \\
mixup & 0.8  \\
cutmix & 1.0 \\
drop path & {0.2} \\
dropout & 0.0 \\
layer-wise lr decay & 0.7  \\
test clips & 5 \\
test crops & 3\\
\bottomrule
\end{tabular}
\end{center}
% \vspace{.2em}
\caption{\textbf{Fine-tuning settings (ViT-B)}}
\label{tab:Finetune} 
\end{table}

\begin{table}[h]
% \tablestyle{1pt}{1.02}
\begin{center}
\begin{tabular}{c|c}
\toprule
Hyperparameter & Value \\
\midrule
training epochs & 100 \\
optimizer & {AdamW} \\ 
base learning & {1e-2}\\
weight decay & {0.0} \\
\bottomrule
\end{tabular}
\end{center}
% \vspace{.2em}
\caption{\textbf{Linear probing settings (ViT-B)}}
\label{tab:Linear Probe} 
\end{table}

\begin{table}[h]
% \tablestyle{1pt}{1.02}
\begin{center}
\begin{tabular}{c|c}
\toprule
Hyperparameter & Value \\
\midrule
Initial speed range & (1.2, 3.0) \\
Acceleration speed range & (-0.06, 0.06) \\ 
Rotation speed range & ($-\frac{1}{100}\pi$, $\frac{1}{100}\pi$)\\
Scale X speed range & (-0.005,0.005) \\
Scale Y speed range & (-0.005,0.005) \\
Shear X speed range & (-0.005,0.005) \\
Shear Y speed range & (-0.005,0.005) \\
\bottomrule
\end{tabular}
\end{center}
\caption{\textbf{Dataset generation settings}}
\label{tab:Dataset} 
\end{table}

\subsection{Additional generated datasets}
\label{app:additional_datasets}
Apart from the progressions presented in the main paper,  we explored video dataset properties from other perspectives. including but not limited to object dynamics, texture information, frame diversity, and real data usage. We provide a brief description of additional datasets below.  
\begin{itemize}
    \item \textbf{Moving objects with slower speed}: For this family of datasets, We repeat some of the progressions mentioned in the main paper, but with slower movement (50\% of the speed of the main progression) and study how the velocity affects the temporal information. The datasets used for this setting include moving circles, moving shapes, moving and transforming shapes, accelerating transforming shapes, and accelerating transforming textures. We present the results of datasets with slower dynamics in \Cref{tab:additional_dynamic}.

    \item \textbf{More texture types}: As discussed in \cref{sec:sub:textures}, we studied different textures settings. In addition to the results in \cref{tab:textures_analysis} and \cref{tab:imagenet_analysis}, we generated some other textures-related data for better understanding. The results are shown in \cref{tab:additional_textures}
    \begin{itemize}
        \item \textbf{Dynamic StyleGAN high-freq}: A less diverse StyleGAN generator from \cite{baradad2021learning}, with only high frequency noise as input to build image structure. We make a new dataset from it by gradually adding random noise as mentioned in the main paper.
        \item \textbf{Replacing with statistic videos from StyleGAN}: Same as the last setting in \cref{sec:sub:imagenet}, we also replace 5\% of the accelerating transforming shapes into StyleGAN samples, which are repeated 16 times to mimic a video.
        \item \textbf{150k images and 150k statistical textures}: we create a dataset that incorporates crops from half of the images and crops from half of the statistical textures we used in the previous dataset in the progression. We apply the same operation in this dataset as in the main progression.
    \end{itemize}

    \item \textbf{More diverse background}: To introduce more diversity into the video, we try to replace the default black background with more diverse and semantic meaningful images. The results are shown in \cref{tab:additional_textures}
    \begin{itemize} 
    \item \textbf{Image crops, with colored background}: We took the same generation setting from '300k images' in the \cref{tab:imagenet_analysis}. For each video, instead of a black background, we randomly sample a color and use it as a background.
    \item \textbf{Image crops, with image background}: Same as the setting above, except that we use a random image from the image crops set to serve as background in each video. 
    \end{itemize}

    \item \textbf{Real data mixture}: Given the powerful ability of synthetic data, we aim to find out if real data and synthetic data can boost each other in downstream tasks. The accuracy for the UCF101 fine-tune setting is presented in \cref{tab:additional_mix_real}.
    \begin{itemize}
        \item \textbf{Accelerating and transforming textures, mix with real video data}: We try replacing 25\% and 75\% of training data by sampling real videos from the UCF101 training set.
        \item \textbf{50\% imagenet crops and 50\% UCF101}: We create a new dataset by mixing samples from the last dataset in the progression in the main paper, and the UCF101 dataset. We make sure that the sample rate is 1:1 and the training size is the same as standard experiments. 
    \end{itemize}

    \item \textbf{Saturated textures}: During exploration, we create a different set of textures-based datasets by making a saturated color version of the datasets. For each moving object, we sampled a random color and added it to the texture crops. Surprisingly, despite the possible corruption in the texture information, they still present competitive performance. A full list of color-saturated datasets and present the results in \cref{tab:additional_saturated}

\end{itemize}

\begin{table}[!htbp]
  \centering
  \begin{tabular}{l|c}
    \toprule
    Dataset configuration &  UCF101\\
    \midrule
    Moving circles  & 84.9\\
    Moving shapes & 88.3\\
    Moving and transforming shapes & 88.3\\
    Accelerating and transforming shapes & 88.6\\
    Accelerating and transforming textures  & 90.9 \\
    \bottomrule
  \end{tabular}
  \caption{\textbf{Additional datasets (ViT-B).} Moving objects with slower speed}
  \label{tab:additional_dynamic}
\end{table}

\begin{table}[!htbp]
  \centering
  \begin{tabular}{l|c}
    \toprule
    Dataset configuration &  UCF101\\
    \midrule
    Dynamic StylaGAN high-frequency  & 68.7\\
    Replacing 5\% of videos w/ StyleGAN  & 88.2\\
    150k images \& 150k statistical textures & 89.7\\
    300k images w/ colored background & 89.9 \\
    300k images w/ image background & 91.0 \\
    \bottomrule
  \end{tabular}
  \caption{\textbf{Additional datasets (ViT-B).} More texture types and more diverse backgrounds}
  \label{tab:additional_textures}
\end{table}

\begin{table}[!htbp]
  \centering
  \begin{tabular}{l|c}
    \toprule
    Dataset configuration &  UCF101\\
    \midrule
     Shapes, 25\% w/ UCF101 & 90.4\\
    Shapes, 75\% w/ UCF101 & 90.6\\
    Image crops, 50\% w/ UCF101 & 92.0 \\
    \bottomrule
  \end{tabular}
  \caption{\textbf{Additional datasets (ViT-B) mixed with with real videos}. Accelerating and transforming shapes and image crops, mixed with real videos with mixing rates.}
  \label{tab:additional_mix_real}
\end{table}

\begin{table}[h]
  \centering
  \begin{tabular}{l|c}
    \toprule
    Dataset configuration &  UCF101\\
    \midrule
    Statistical textures & 88.9\\
    Statistical textures w/ colored background & 87.8\\
    Moving Dynamic StyleGAN crops & 87.5\\
    300k image crops & 90.1\\
    150k image crops \& 150 statistical textures & 89.2\\
    300k image crops w/ colored background  & 89.5\\
    300k image crops w/ image background & 89.5\\
    1.3M image crops & 89.8\\
    \bottomrule
  \end{tabular}
  \caption{\textbf{Additional datasets (ViT-B) - saturated textures.}}
  \label{tab:additional_saturated}
\end{table}

\end{document}